\documentclass[graybox]{svmult}

\usepackage{mathptmx}       
\usepackage{helvet}         
\usepackage{courier}        
\usepackage{type1cm}        
\usepackage{makeidx}         
\usepackage{graphicx}        
\usepackage{multicol}        
\usepackage[bottom]{footmisc}

\usepackage[numbers]{natbib}
\usepackage{multicol}
\usepackage[bookmarks=true]{hyperref}

\usepackage[font={scriptsize}]{caption}

\usepackage{bbm}

\usepackage{graphicx} 

\usepackage{amssymb}
\usepackage{gensymb}

\usepackage{amsmath}
\usepackage{wrapfig}
\usepackage{color}
\usepackage{dsfont}
\usepackage{mathrsfs}

\usepackage{soul}

\usepackage{xr}
\newcommand{\todo}[1]{\textcolor{blue}{[\textbf{#1}]}}

\newcommand{\argmin}{\arg\!\min}
\newcommand{\argmax}{\arg\!\max}

\newcommand{\jf}[1]{#1}

\newcommand{\jae}[1]{#1}
\newcommand{\jin}[1]{#1}

\makeindex             

\begin{document}

\title*{Robobarista: Object Part based Transfer of\\ Manipulation Trajectories from Crowd-sourcing in 3D Pointclouds}
\titlerunning{Robobarista: Object Part based Transfer of Manipulation Trajectories}

\author{Jaeyong Sung, Seok Hyun Jin, and Ashutosh Saxena}
\institute{Jaeyong Sung, Seok Hyun Jin, and Ashutosh Saxena 
\at Department of Computer Science, Cornell University, USA. \\
e-mail: \texttt{\{jysung,sj372,asaxena\}@cs.cornell.edu}
}
%
%
\maketitle

\newcommand{\abstracttext}{
There is a large variety of objects and appliances in human environments, such as stoves, 
coffee dispensers, juice extractors, and so on.  It is challenging for a roboticist to program a 
robot for each of these object types and for each of their instantiations.
In this work, we present a novel approach to manipulation planning
based on the idea that many household objects share similarly-operated object parts.
We formulate the manipulation planning as a structured prediction problem
and design a deep learning model that can handle large noise in the manipulation demonstrations
and learns features from three different modalities: point-clouds, language and trajectory.
In order to collect a large number of manipulation demonstrations for different objects,
we developed a new crowd-sourcing platform called Robobarista.
We test our model on our dataset consisting of 116 objects with 249 parts
along with 250 language instructions, for which there are 1225 crowd-sourced 
manipulation demonstrations.
We further show that our robot can even manipulate objects it has never
seen before.
}

\abstract*{\abstracttext}
\abstract{\abstracttext}

\noindent \textbf{\textit{Keywords---}}  \textbf{Robotics and Learning}, Crowd-sourcing, Manipulation


\abovedisplayskip 3.0pt plus2pt minus2pt
\belowdisplayskip \abovedisplayskip
 \renewcommand{\baselinestretch}{0.99} 

\newenvironment{packed_enum}{
\begin{enumerate}
  \setlength{\itemsep}{0pt}
  \setlength{\parskip}{0pt}
  \setlength{\parsep}{0pt}
}
{\end{enumerate}}

\newenvironment{packed_item}{
\begin{itemize}
  \setlength{\itemsep}{0pt}
  \setlength{\parskip}{0pt}
  \setlength{\parsep}{0pt}
}{\end{itemize}}

\newlength{\sectionReduceTop}
\newlength{\sectionReduceBot}
\newlength{\subsectionReduceTop}
\newlength{\subsectionReduceBot}
\newlength{\abstractReduceTop}
\newlength{\abstractReduceBot}
\newlength{\captionReduceTop}
\newlength{\captionReduceBot}
\newlength{\subsubsectionReduceTop}
\newlength{\subsubsectionReduceBot}

\newlength{\horSkip}
\newlength{\verSkip}

\newlength{\figureHeight}
\setlength{\figureHeight}{1.7in}

\setlength{\horSkip}{-.09in}
\setlength{\verSkip}{-.1in}
\setlength{\subsectionReduceTop}{-0.33in}
\setlength{\subsectionReduceBot}{-0.18in}
\setlength{\sectionReduceTop}{-0.27in}
\setlength{\sectionReduceBot}{-0.17in}
\setlength{\subsubsectionReduceTop}{-0.06in}
\setlength{\subsubsectionReduceBot}{-0.05in}
\setlength{\abstractReduceTop}{-0.05in}
\setlength{\abstractReduceBot}{-0.10in}

\setlength{\captionReduceTop}{-0.20in}
\setlength{\captionReduceBot}{-0.17in}


\vspace*{\sectionReduceTop}
\section{Introduction}
\vspace*{\sectionReduceBot}

\begin{wrapfigure}{r}{0.6\textwidth}

  \begin{center}
    \vskip -.12in
    \includegraphics[width=0.55\textwidth]{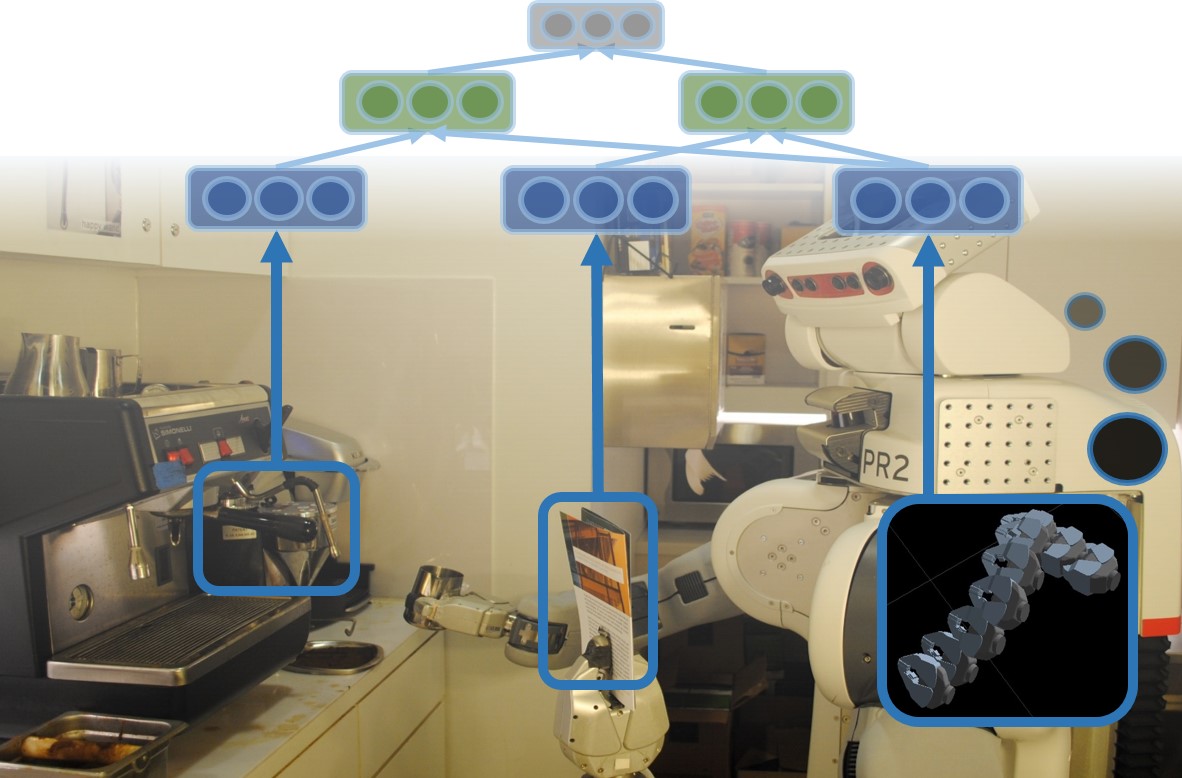}
  \end{center}
  \vskip -.2in
  \caption{ 
  \textbf{First encounter of an espresso machine} by our PR2 robot. Without ever having seen the machine before,  given the language instructions and a point-cloud from Kinect sensor, our robot is 
  capable of finding appropriate   manipulation trajectories from prior experience using our deep learning model.}
  \label{fig:robobarista_main}
\vspace{-.12in}
\end{wrapfigure}

Consider the espresso machine in Figure~\ref{fig:robobarista_main} ---
even without having seen the machine before, a person can prepare a cup of latte
by visually observing the machine and
by reading a natural language instruction manual. 
This is possible because humans have vast prior experience of manipulating
differently-shaped objects that share common parts such as `handles' and `knobs'.
\jin{In this work, our goal is to enable robots to generalize their manipulation ability to novel objects and tasks} (e.g. toaster, sink, water fountain, toilet, soda dispenser).
Using a large knowledge base of manipulation demonstrations,
we build an algorithm that infers \jin{an appropriate} manipulation trajectory 
given a point-cloud and natural language instructions.


\begin{figure*}[tb]
\begin{center}
\vskip -.06in
\includegraphics[width=\textwidth,trim={3mm 0 1mm 0}]{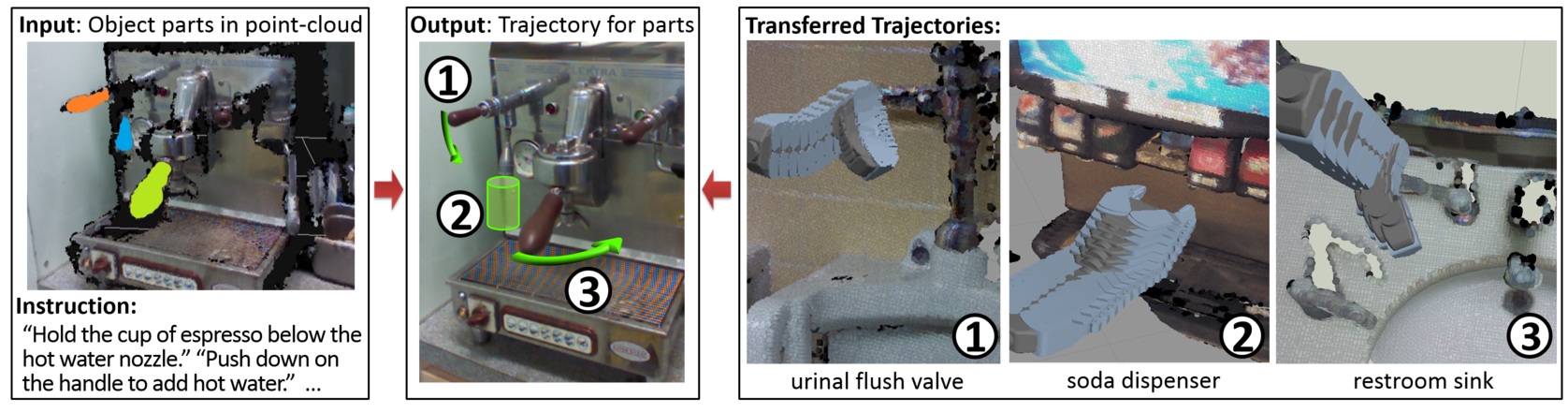}
\end{center}
\vskip -.23in
\caption{\textbf{Object part and natural language instructions input to manipulation trajectory as output.} Objects such as the espresso machine
consist of distinct object parts, each of which requires a distinct manipulation trajectory for manipulation. 
For each part of the machine, we can re-use a manipulation trajectory that was used for some other object with 
similar parts. So, for an object part in a point-cloud (each object part colored on left), we 
can find a trajectory used to manipulate some other object (labeled on the right) that can be 
\emph{transferred} (labeled in the center). With this approach, a robot can operate a new and previously 
unobserved object such as the `espresso machine', 
by successfully transferring trajectories from other completely different but previously observed objects.
Note that the input point-cloud is very noisy and incomplete (black represents missing points).}
\label{fig:espresso_transfer}
\vskip -.2in
\end{figure*}

The key idea in our work is that many objects designed for humans
share many similarly-operated \emph{object parts} such as `handles', `levers', `triggers', and `buttons';
and manipulation motions can be transferred even among completely different objects
if we represent motions  with respect to \textit{object parts}.
For example, even if the robot has never seen the `espresso machine' before, the robot should be able to
manipulate it if it has previously seen similarly-operated parts 
in other objects such as `urinal', `soda dispenser', and `restroom sink' 
as illustrated in Figure~\ref{fig:espresso_transfer}.
Object parts that are operated in similar fashion may not carry the same part name (e.g., `handle')
but would rather have some similarity in their shapes \jin{that \jf{allows}} the motion to be transferred
between completely different objects.

If the sole task for the robot is to manipulate one specific espresso machine or 
just a few types of `handles', 
a roboticist could manually program the exact sequence to be executed.
However, in human environments, there is 
a large variety in the types of object and their instances. 
\jae{Classification} of objects or object parts (e.g. `handle') alone does not provide enough information for robots to
actually manipulate them.
Thus, rather than relying on scene understanding techniques \cite{blaschko2008learning,li2009towards,girshick2011object},
we directly use 3D point-cloud for manipulation planning
using machine learning algorithms.

\jin{
Such machine learning algorithms require a large dataset for training. 
However, collecting such large dataset of expert demonstrations is very expensive as 
it requires joint physical presence of the robot, an expert, and the object to be manipulated.
In this work, we show that we can crowd-source the collection of manipulation demonstrations to the
public over the web through our Robobarista  platform and still outperform the model trained with
expert demonstrations.
}

\jin{The key challenges in our problem  are in designing features and a learning model 
that integrates three completely different modalities of data (point-cloud, language and trajectory),
and in  handling significant amount of noise in crowd-sourced manipulation demonstrations.
Deep learning has made impact in related application areas 
(e.g., vision \cite{krizhevsky2012imagenet,bengio2013representation}, natural language processing \cite{socher2011semi}).
In this work, we present a deep learning model that can handle large noise in labels,
with a new architecture that learns relations between the three different modalities.}
\jin{Furthermore, in contrast to previous approaches
based on
learning from demonstration (LfD) that learn a mapping from a state to
 an action \cite{argall2009survey},
our work complements LfD as we focus on the entire manipulation motion
(as opposed to a sequential state-action mapping).}

In order to validate our approach, we have collected a large dataset of \emph{116 objects} 
with \emph{250 natural language instructions}
for which there are \emph{1225 crowd-sourced manipulation trajectories} from 71 non-expert users 
via our Robobarista web platform ({\small \url{http://robobarista.cs.cornell.edu}}).
We also present experiments on our robot using our approach.
In summary, the key contributions of this work are:
\begin{itemize}
\item a novel approach to manipulation planning via \textit{part-based transfer} between
different objects \jin{that allows manipulation of novel objects},
\item incorporation of \textit{crowd-sourcing} to manipulation planning, 
\item introduction of \textit{deep learning model} that handles three modalities with noisy labels from crowd-sourcing, and
\item contribution of the first large manipulation dataset and experimental evaluation on this dataset.
\end{itemize}

\vspace*{\sectionReduceTop}
\section{Related Work}
\vspace*{\sectionReduceBot}

\noindent
\textbf{Scene Understanding.}
There has been great advancement in scene understanding \cite{li2009towards,koppula2011semantic,wu2014_hierarchicalrgbd},
in human activity detection \cite{sung_rgbdactivity_2012,Hu2014humanact},
and in features for RGB-D images and point-clouds \cite{socher2012convolutional,lai_icra14}.
And, similar to our idea of using part-based transfers, 
the deformable part model \cite{girshick2011object} was effective
in object detection.
\jae{However, classification of objects, object parts, or human activities alone
does not provide enough information for a robot to reliably plan manipulation.}
\jin{Even a simple category such as ‘kitchen sinks’ has so much variation in its instances, 
each differing in how it is operated: 
pulling the handle upwards, pushing upwards, pushing sideways, and so on.}
On the other hand, direct perception approach skips the intermediate object labels
and directly perceives affordance based on the shape of the object
\cite{gibson1986ecological,kroemer2012kernel}.
It focuses on detecting the part known to afford certain action
such as `pour' given the object,
while we focus on predicting the correct motion given the object part.

\noindent
\textbf{Manipulation Strategy.}
For highly \jin{specific} tasks,
many works manually sequence different controllers
to accomplish complicated tasks such as baking cookies \cite{bollini2011bakebot}
and folding the laundry \cite{miller2012geometric},
or focus on learning specific motions such as grasping \cite{kehoe2013cloud}
and opening doors \cite{endres2013learning}.
\jin{Others focus on learning to sequence different movements
\cite{sung_synthesizingsequences_2014, misra2014tellme} 
but assume that there exist perfect controllers such as \textit{grasp} and \textit{pour}.}

For a more general task of manipulating new instances of objects, previous approaches rely 
on finding articulation \cite{sturm2011probabilistic,pillai2014articulated} 
or using interaction \cite{katz2013interactive}, but they are limited by tracking performance of
a vision algorithm.
Many objects that humans operate daily have parts such as ``knob'' that are small, which leads to
significant occlusion as manipulation is demonstrated.
Another approach using part-based transfer between objects has been shown to be successful
for grasping \cite{dang2012semantic,detry2013learning}.
We extend this approach and introduce a deep learning model that
enables part-based transfer of \jin{\textit{trajectories}}
by automatically learning relevant features.
Our focus is on the generalization of manipulation trajectory via part-based transfer
using point-clouds without knowing objects a priori
and without assuming any of the sub-steps (`approach', `grasping', and `manipulation').

\noindent
\textbf{Learning from Demonstration (LfD).}
The most successful approach for teaching robots \jae{tasks},
such as helicopter maneuvers \cite{abbeel2010autonomous}
or table tennis \cite{mulling2013learning},
has been based on LfD \cite{argall2009survey}.
Although LfD allows end users to demonstrate the task by simply taking the robot arms,
it focuses on learning individual actions and separately relies on high level task composition 
\cite{mangin2011unsupervised,daniel2012learning}
or is often limited to previously seen objects \cite{phillipslearning, pastor2009learning}.
\jin{We} believe that learning a single model for an action like ``turning on''
is impossible because human environment has many variations.

Unlike learning a model from demonstration,
instance-based learning \cite{aha1991instance,forbes2014robot} replicates
one of the demonstrations.
Similarly, we directly transfer one of the demonstrations but
focus on generalizing manipulation planning to completely new objects,
enabling robots to manipulate objects they have never seen before.

\noindent
\textbf{Deep Learning.}
There has been great success with deep learning, especially in the domains
of vision and natural language processing (e.g. \cite{krizhevsky2012imagenet,socher2011semi}).
In robotics, deep learning has previously  been  successfully used
for detecting grasps on multi-channel input of RGB-D images \cite{lenz2013deep}
and for classifying terrain from long-range vision \cite{hadsell2008deep}.

Deep learning can also solve multi-modal problems \cite{ngiam2011multimodal,lenz2013deep}
and structured problems \cite{socher2011parsing}.
Our work builds on prior works and extends neural network
to handle three modalities which are of completely different data type (point-cloud, language, and trajectory)
while handling lots of label-noise originating from crowd-sourcing.

\noindent
\textbf{Crowd-sourcing.}
Teaching robots how to manipulate different objects has often relied on experts
\cite{argall2009survey,abbeel2010autonomous}.
Among previous efforts to scale teaching to the crowd \cite{crick2011human,tellex2014asking,jainsaxena2015_learningpreferencesmanipulation},
\citet{forbes2014robot} employs a similar approach towards crowd-sourcing
but collects
multiple instances of similar table-top manipulation with same object,
and others build web-based platform for crowd-sourcing manipulation \cite{toris2013robotsfor,toris2014robot}.
These approaches either depend on the presence of an expert (due to a required 
special software)
or require a real robot at a remote location.
Our Robobarista platform borrows some components of \cite{alexander2012robot},
but works on any standard web browser with OpenGL support and incorporates real point-clouds of various scenes.

\vspace*{\sectionReduceTop}
\section{Our Approach}
\vspace*{\sectionReduceBot}
The intuition for our approach is that
many differently-shaped objects share similarly-operated object parts; 
thus, the manipulation trajectory of an object can be transferred to a
completely different object if they share similarly-operated parts.
We formulate this problem as a structured prediction problem
and introduce a deep learning model
that handles three modalities of data 
and deals with noise in crowd-sourced data.
Then, we introduce the crowd-sourcing platform Robobarista 
to easily scale the collection of manipulation demonstrations to 
\jae{non-experts} on the web.

%



\vspace*{\subsectionReduceTop}
\subsection{Problem Formulation}
\vspace*{\subsectionReduceBot}

\label{sec:prob_form}
The goal is to learn a function $f$ that maps a given pair of point-cloud 
$p \in \mathcal{P}$ of object part
and language $l \in \mathcal{L}$ to a trajectory $\tau \in 
\mathcal{T}$ that can manipulate the object part as described by \jae{free-form natural language} $l$:
$$f : \mathcal{P} \times \mathcal{L} \rightarrow \mathcal{T}$$

\noindent
\textbf{Point-cloud Representation.}
Each instance of point-cloud $p \in \mathcal{P}$ is represented as a set of $n$ points in three-dimensional
Euclidean space where each point $(x,y,z)$ is represented with its RGB color $(r,g,b)$:
$p = \{p^{(i)} \}^n_{i=1} = {\{(x,y,z,r,g,b)^{(i)}\}}^n_{i=1}$.
The size of the set vary \jin{for each instance}.
These points are often obtained by stitching together a sequence of sensor data 
from an RGBD sensor \cite{izadi2011kinectfusion}.

\noindent
\textbf{Trajectory Representation.}
\label{sec:trajrep}
Each trajectory $\tau \in \mathcal{T}$ is represented as a sequence of $m$ \textit{waypoints},
where each waypoint consists of gripper status $g$, translation $(t_x, t_y, t_z)$,
and rotation $(r_x, r_y, r_z, r_w)$ with respect to the origin:
$\tau= \{\tau^{(i)}\}^m_{i=1} = {\{(g, t_x, t_y, t_z, r_x, r_y, r_z, r_w)^{(i)}\}}^m_{i=1}$
where  $g \in \{\text{``open''},\text{``closed''},\text{``holding''}\}$.
$g$ depends on the type of the end-effector,
which we have assumed to be a two-fingered gripper like that of PR2 or Baxter.
The rotation is represented as quaternions $(r_x, r_y, r_z, r_w)$
instead of the more compact Euler angles to prevent problems such as
the gimbal lock \cite{saxena2009learning}.

\noindent
\textbf{Smooth Trajectory.} To acquire a smooth trajectory from
a waypoint-based trajectory $\tau$, 
we interpolate intermediate waypoints.
Translation is linearly interpolated and the quaternion is interpolated
using spherical linear interpolation (Slerp) \cite{shoemake1985animating}.

\vspace*{\subsectionReduceTop}
\subsection{Can transferred trajectories adapt without modification?}
\label{sec:transfer_adapt}
\vspace*{\subsectionReduceBot}


Even if we have a trajectory to transfer,
a conceptually transferable trajectory is not necessarily directly compatible
if it is represented with respect to an inconsistent reference point.

\jin{To make a trajectory compatible with a new situation without modifying the 
trajectory,
we need a representation method for trajectories, based on point-cloud information,
that allows a \textit{direct
transfer of a trajectory without any modification}.}

\noindent
\textbf{Challenges.}
Making a trajectory compatible when transferred to a different object or to a
different instance of the same object without modification
 can be challenging depending on the representation
of trajectories and the variations in the location of the object, 
given in point-clouds. 


For robots with high degrees of freedom arms such as PR2 or Baxter robots,
trajectories are commonly represented as a sequence of joint angles (in configuration space)
\cite{thrun2005probabilistic}.
With such representation, the robot needs to modify the trajectory for an object
with forward and inverse kinematics even
for a small change in the object's position and orientation.
Thus, trajectories in the configuration space 
are prone to errors as they are realigned with the object.
They can be executed without modification
 only when the robot is in the exact same position and orientation
with respect to the object.

One approach that allows execution without modification 
is representing trajectories with respect to the object
by aligning via point-cloud registration (e.g. \cite{forbes2014robot}). 
However, if the object is large (e.g. a stove) and has many parts (e.g. knobs 
and handles), then object-based
representation is prone to errors when individual parts have different
translation and rotation.
This limits the transfers to be between different instances of the
same object that is small or has a simple structure.

\jin{
Lastly, it is even more challenging if two objects require similar trajectories, but have \jae{slightly} different shapes.
And this is made more difficult by limitations of the point-cloud data. 
As shown in left of Fig.~\ref{fig:espresso_transfer},
the point-cloud data, even when stitched from multiple angles, are very noisy 
compared to the RGB images. 
}

\noindent
\textbf{Our Solution.}
Transferred trajectories become compatible across different objects when trajectories are 
represented
1) in the task space rather than the configuration space, and 
2) in the principal-axis based coordinate frame of the object \textit{part} rather than the robot or the object.

Trajectories can be represented in the task space by recording only the position and orientation
of the end-effector.
By doing so, we can focus on the actual interaction between the robot and the 
environment
rather than the movement of the arm.
It is very rare that the arm configuration affects the completion of the task as long as there is no collision.
With the trajectory represented as a sequence of gripper position and orientation, 
the robot can find its arm configuration that 
is collision free with the environment using inverse kinematics.

However, representing the trajectory in task space is not enough to make transfers
compatible.
It has to be in a common coordinate frame regardless of object's orientation and shape. 
Thus, we align the negative $z$-axis along gravity and align the $x$-axis along the
principal axis of the object \textit{part} using PCA \cite{hsiao2010contact}.
With this representation, even when the object part's position and orientation changes,
the trajectory does not need to change.
The underlying assumption is that similarly operated object parts share 
similar shapes leading to \jin{a similar direction in their principal axes}.


\vspace*{\sectionReduceTop}
\section{Deep Learning for Manipulation Trajectory Transfer}
\label{sec:deep}
\vspace*{\sectionReduceBot}

We use deep learning  to find the most appropriate trajectory
for the given point-cloud and natural language.
Deep learning is mostly used for binary or multi-class
classification or regression problem \cite{bengio2013representation} with a uni-modal 
input.
We introduce a deep learning model 
that can handle three completely different modalities of point-cloud, language, and trajectory 
and solve a structural problem with lots of label noise.


The original structured prediction problem
($f : \mathcal{P} \times \mathcal{L} \rightarrow \mathcal{T}$)
is converted to a binary classification problem 
($f : (\mathcal{P} \times \mathcal{L}) \times \mathcal{T} \rightarrow \{0, 1\}$).
Intuitively, the model takes the input of point-cloud, language, and trajectory
and outputs whether it is
a good match (label $y=1$) or a bad match (label $y=0$).

\begin{wrapfigure}{r}{0.53\textwidth}
\begin{center}
\vskip -0.3in
\includegraphics[width=0.5\textwidth]{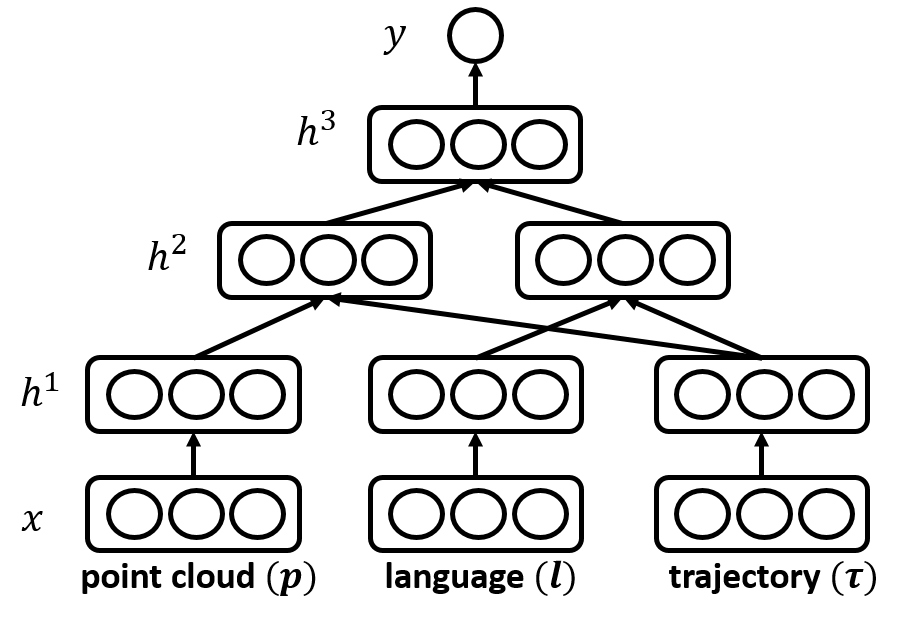}
\end{center}
\vskip -0.2in
\caption{
\scriptsize
\textbf{Our deep learning model} for transferring manipulation trajectory.
Our model takes the input $x$ of three different modalities (point-cloud, language, and trajectory)
and outputs $y$, whether it is a good match or bad match. It first learns features separately ($h^1$) for each modality 
and then \jf{learns the} relation ($h^2$) between input and output of the original structured problem.
Finally, last hidden layer $h^3$ learns relations of all these modalities.
}
\label{fig:deep}
\vskip -0.14in
\end{wrapfigure}

\noindent
\textbf{Model.}
Given an input of point-cloud, language, and trajectory, $x=((p,l),\tau)$,
as shown at the bottom of Figure~\ref{fig:deep},
the goal is to classify as either $y=0$ or $1$ at the top.
The first $h^1$ layer learns a separate layer of features for each modality of $x\; (=h^0)$ \cite{ngiam2011multimodal}.
The next layer learns the relations between the input $(p,l)$
and the output $\tau$ of the original structured problem, 
combining two modalities
at a time. The left combines point-cloud and trajectory
and the right combines language and trajectory.
The third layer $h^3$ learns the relation
between these two combinations of modalities and 
the final layer $y$ represents the binary label. 

Every layer $h^i$ uses the rectified linear unit \cite{zeiler2013rectified}
as the activation function: 
$$h^i = a(W^i h^{i-1} + b^i) \text{ where } a(\cdot)=max(0,\cdot)$$
with weights to be learned
$W^i \in \mathds{R}^{M \times N}$, where $M$ and $N$ represent the number of nodes in $(i-1)$-th and $i$-th layer respectively. 
The logistic regression is used in last layer for predicting 
the final label $y$.
The probability that $x = ((p,l),\tau)$ is a ``good match'' is computed as: 
$P(Y=1|x;W,b) = 1/(1 + e^{-(Wx+b)})$

\noindent
\textbf{Label Noise.}
When data contains lots of noisy label (noisy trajectory $\tau$) due to crowd-sourcing, not all
crowd-sourced trajectories should be trusted as equally appropriate as will be shown
in Sec.~\ref{sec:experiments}.

For every pair of input $(p,l)_i$, we have 
$\mathcal{T}_i=\{\tau_{i,1}, \tau_{i,2}, ..., \tau_{i,n_i} \}$, 
a set of trajectories submitted by the crowd for $(p,l)_i$. 
First, the best candidate label $\tau^*_i \in \mathcal{T}_i$ for $(p,l)_i$ is selected as one of the labels with the
smallest average trajectory distance (Sec.~\ref{sec:metric}) to other labels:
$$\tau^*_i = \argmin_{\tau \in \mathcal{T}_i} \frac{1}{n_i} \sum_{j=1}^{n_i} \Delta(\tau, \tau_{i,j}) $$
We assume that at least half of the crowd tried to give a reasonable demonstration.
Thus a demonstration with the smallest average distance to all other demonstrations
must be a good demonstration.

\jae{Once} we have found the most likely label
$\tau^*_i$ for $(p, l)_i$, we give the label 1 (``good match'') to
$((p, l)_i, \tau^*_i)$, making it the first positive example for the binary
classification problem. Then we find more positive examples by finding 
other trajectories $\tau' \in \mathcal{T}$ 
such that $\Delta(\tau^*_i,\tau') < t_g$ where $t_g$ is a threshold determined by the expert.
Similarly, negative examples are generated by finding trajectories $\tau' \in \mathcal{T}$ 
such that it is above some threshold $\Delta(\tau^*_i,\tau') > t_w$,
where $t_w$ is determined by expert, and they are given label 0 (``bad match'').


\noindent
\textbf{Pre-training.} 
We use the stacked sparse de-noising auto-encoder (SSDA) to train weights $W^i$ and bias $b^i$  for each layer
\cite{vincent2008extracting,zeiler2013rectified}.
Training occurs layer by layer from bottom to top
trying to reconstruct the previous layer using SSDA.
To learn parameters for layer $i$, we build an auto-encoder
which takes the corrupted output $\tilde{h}^{i-1}$ (binomial noise with corruption level $p$) of previous layer as input
and minimizes the loss function \cite{zeiler2013rectified} with max-norm constraint \cite{srivastava2013improving}:

{
\vskip -.15in
\small
\begin{align*}
&  W^* = \argmin_W \lVert \hat{h}^{i-1} - h^{i-1} \rVert^2_2 + \lambda \lVert h^{i} \rVert_1 \\
&\text{where}\quad  \hat{h}^{i-1}  = f(W^i h^i + b^i) \qquad  h^i = f({W^i}^T \tilde{h}^{i-1} + b^i)  \qquad \quad \;\;  \tilde{h}^{i-1} = h^{i-1} X   \\
&  \qquad\quad\;\;  \lVert W^i \rVert_2 \leq c  \qquad \qquad \quad \; \;\;\;  X \sim  B(1,p)  
\end{align*}
\vskip -.05in
}

\noindent
\textbf{Fine-tuning.}
The pre-trained neural network can be fine-tuned by minimizing the negative log-likelihood 
with the stochastic gradient method with mini-batches:
$NLL = - \sum_{i=0}^{|D|} log(P(Y=y^{i}|x^{i},W,b))$.
To prevent over-fitting to the training data, we used dropout 
\cite{hinton2012improving}, which randomly drops a specified percentage of the output
of every layer.

\noindent
\textbf{Inference.} 
Given the trained neural network, inference step finds the trajectory $\tau$
that maximizes the output through sampling in the space of trajectory $\mathcal{T}$: 
$$\argmax_{\tau' \in \mathcal{T}} P(Y=1|x=((p,l),\tau');W,b)$$
Since the space of trajectory $\mathcal{T}$ is infinitely large,
based on our idea that we can transfer trajectories across objects,
we only search trajectories that the model has seen in training phase.

\noindent
\textbf{Data pre-processing.}
As seen in Sec.~\ref{sec:prob_form},
each of the modalities $(p, l, \tau)$ can have any length. 
Thus, we pre-process to make each fixed in length.

We represent point-cloud $p$ of any arbitrary length
as an occupancy grid where each cell 
indicates whether any point lives in the space it represents.
Because point-cloud $p$ consists of only the part of an object which is limited in size,
we can represent $p$ using two occupancy grids of size $10 \times 10 \times 10$ with different scales: 
one with each cell representing $1 \times 1 \times 1 (cm)$ and the other with 
each cell representing $2.5\times 2.5 \times 2.5 (cm)$.

Each language instruction is represented as a fixed-size bag-of-words representation with
stop words removed.
Finally, for each trajectory $\tau \in \mathcal{T}$, 
we first compute its smooth interpolated trajectory $\tau_s \in \mathcal{T}_s$
(Sec.~\ref{sec:trajrep}),
and then normalize all trajectories $\mathcal{T}_s$ to the same length
while preserving the sequence of gripper status.



\vspace*{\sectionReduceTop}
\section{Loss Function for Manipulation Trajectory}
\label{sec:metric}
\vspace*{\sectionReduceBot}

Prior metrics for trajectories consider only their translations
(e.g. \cite{koppula2013_anticipatingactivities})
and not their rotations \emph{and} gripper status.
\jae{We propose a new measure, which uses dynamic time warping, for evaluating manipulation trajectories.}
This measure non-linearly warps two trajectories of 
arbitrary lengths to produce a matching, and 
cumulative distance is computed \jin{as the} sum of cost of all matched waypoints.
The strength of this measure
is that weak ordering is maintained among matched waypoints and that every waypoint 
contributes to the cumulative distance.

For two trajectories of arbitrary lengths,
$\tau_A = \{\tau_A^{(i)}\}_{i=1}^{m_A}$ and
$\tau_B = \{\tau_B^{(i)}\}_{i=1}^{m_B}$
, we define matrix $D \in \mathbb{R}^{m_A \times m_B}$, 
where $D(i, j)$ is the cumulative distance of an \jae{optimally-warped} matching
between trajectories up to index $i$ and $j$, respectively, of each trajectory.
The first column and the first row of
$D$ is initialized as
$D(i,1) = \sum_{k=1}^{i}c(\tau_A^{(k)}, \tau_B^{(1)})\, \forall i \in [1, m_A]$ and 
$D(1,j) = \sum_{k=1}^{j}c(\tau_A^{(1)}, \tau_B^{(k)})\, \forall j \in [1, m_B]$, 
where $c$ is a local cost function between two waypoints (discussed later). The rest of $D$
is completed using dynamic programming:
$D(i,j) = c(\tau_A^{(i)}, \tau_B^{(j)}) +  \min\{D(i-1, j-1), D(i-1,j), D(i, j-1)\}$


Given the constraint that $\tau_A^{(1)}$ is matched to $\tau_B^{(1)}$,
the formulation ensures that every waypoint contributes to the
final cumulative distance $D(m_A, m_B)$. Also,
given a matched pair $(\tau_A^{(i)}, \tau_B^{(j)})$, no waypoint
preceding $\tau_A^{(i)}$ is matched to a waypoint succeeding $\tau_B^{(j)}$,
encoding weak ordering. 

The pairwise cost function $c$ between matched waypoints $\tau_A^{(i)}$ and $\tau_B^{(j)}$
is defined:

{
\vskip -.2in
\small
\begin{align*}
c(\tau_A^{(i)}, \tau_B^{(j)};\alpha_T, \alpha_R, \beta, \gamma) =  w(\tau_A^{(i)};\gamma)& w(\tau_B^{(j)};\gamma)  
\bigg(\frac{d_T(\tau_A^{(i)}, \tau_B^{(j)})}{\alpha_T} + \frac{d_R(\tau_A^{(i)}, \tau_B^{(j)})}{\alpha_R}\bigg) \bigg(1 + \beta d_G(\tau_A^{(i)}, \tau_B^{(j)}) \bigg) \\
\text{where} \hspace{5 mm} d_T(\tau_A^{(i)},\tau_B^{(j)}) &= ||(t_x, t_y, t_z)_{A}^{(i)} - (t_x, t_y, t_z)_{B}^{(j)}||_2  \\  
d_R(\tau_A^{(i)},\tau_B^{(j)})  &= \text{angle difference between $\tau_{A}^{(i)}$ and $\tau_B^{(j)}$} \\
\qquad d_G(\tau_A^{(i)},\tau_B^{(j)}) &= \mathbbm{1}(g_A^{(i)}=g_B^{(j)}) \\
w(\tau^{(i)}; \gamma) &= exp(-\gamma \cdot ||\tau^{(i)}||_2) 
\end{align*}
\vskip -.05in
}
\noindent
The parameters $\alpha, \beta$ are for scaling translation and rotation errors, and
gripper status errors, respectively. $\gamma$ weighs the importance of a waypoint based on
its distance \jin{to} the object part. 
Finally, as trajectories vary in length,
we normalize $D(m_A, m_B)$ by the number of waypoint pairs that contribute
to the cumulative sum, $|D(m_A, m_B)|_{path^*}$ 
(i.e. the length of the optimal warping path), giving the final form:
\begin{align*}
distance(\tau_A, \tau_B) = \frac{D(m_A, m_B)}{|D(m_A, m_B)|_{path^*}}
\end{align*}
This distance function is used for noise-handling in our model
and as the final evaluation metric.

\vspace*{\sectionReduceTop}
\section{Robobarista: crowd-sourcing platform}
\vspace*{\sectionReduceBot}

\begin{figure*}[tb]
\begin{center}
\includegraphics[width=\textwidth]{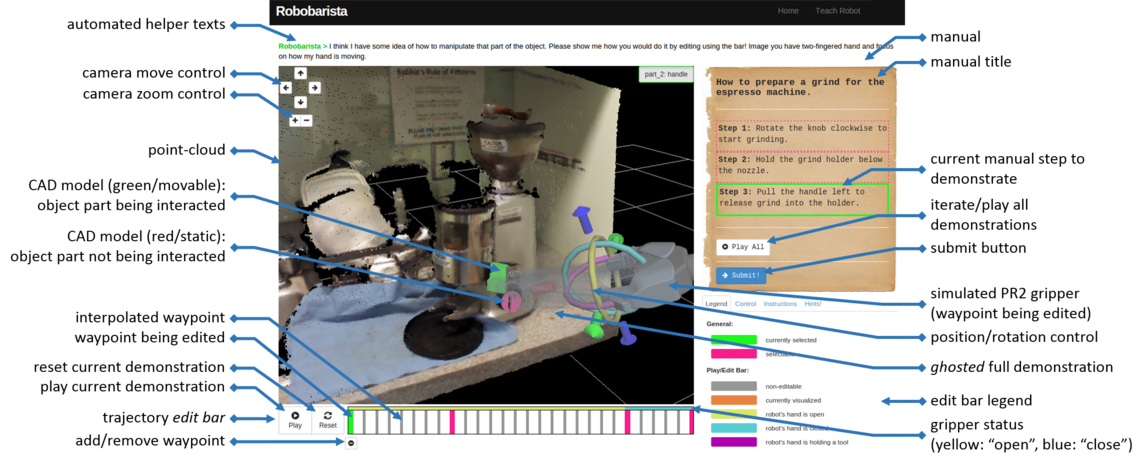}
\end{center}
\vskip -0.2in
\caption{
\textbf{Screen-shot of Robobarista,} the crowd-sourcing platform running on Chrome browser.
We have built Robobarista platform for collecting a large number of crowd \jf{demonstrations} for teaching the robot.
}
\label{fig:robobarista}
\end{figure*}

\begin{figure*}[tb]
\begin{center}
\includegraphics[width=\textwidth]{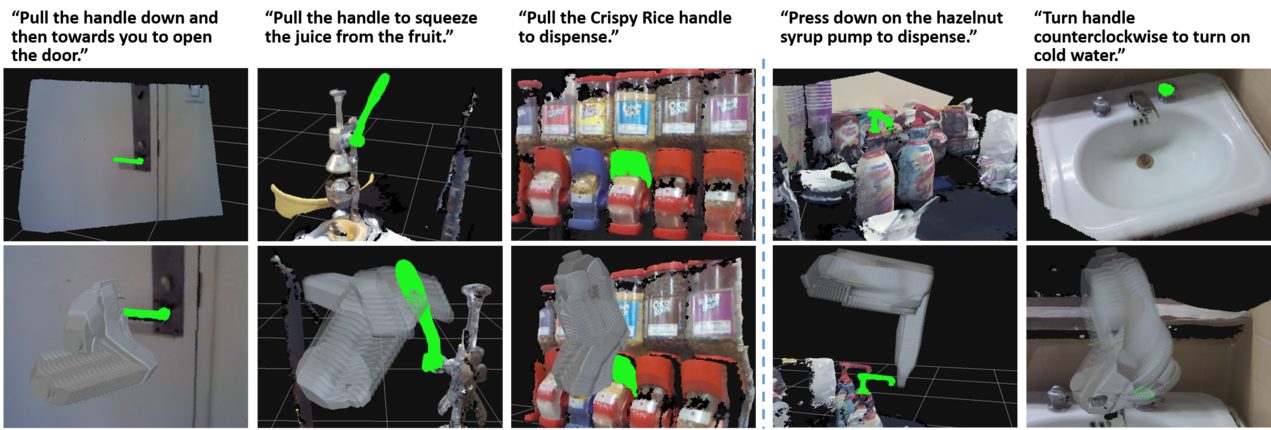}
\end{center}
\vskip -0.2in
\caption{
\textbf{Examples from our dataset,} \jf{each of which  
consists} of a natural language instruction
(top), an object part in point-cloud representation (highlighted), and a manipulation
trajectory (below) collected via Robobarista. 
Objects range from kitchen appliances such as stove and rice cooker to
urinals and sinks in restrooms. As our trajectories are collected from non-experts,
they vary in quality from being likely to complete the manipulation task
successfully (left of dashed line) to being unlikely to do so successfully
(right of dashed line).
}
\label{fig:data_ex}
\vskip -0.1in
\end{figure*}

In order to collect a large number of manipulation demonstrations from the crowd, we built a crowd-sourcing web platform that we call Robobarista (see Fig.~\ref{fig:robobarista}).
It provides a virtual environment where non-expert users can teach robots
via a web browser, without expert guidance or  physical presence with a robot and a target object.

The system simulates a situation where the user encounters 
a previously unseen target object and a natural language instruction manual
for its manipulation.
Within the web browser, users are shown a point-cloud in the 3-D viewer on the left
and a \textit{manual} on the right.
A manual may involve several instructions,
such as ``Push down and pull the handle to open the door''.
The user's goal is to demonstrate how to manipulate
the object in the scene for each instruction.

The user starts by selecting one of the instructions on the right to demonstrate (Fig.~\ref{fig:robobarista}).
Once selected, the target object part
is highlighted and the trajectory \textit{edit bar} appears below the 3-D viewer.
Using the \textit{edit bar}, which works like a video editor, the user can 
playback and edit the demonstration.
Trajectory representation
as a set of waypoints (Sec.~\ref{sec:trajrep}) is directly shown on the \textit{edit bar}.
The bar shows not only the set of waypoints (red/green) but also the interpolated waypoints (gray).
The user can click the `play' button or hover the cursor over the edit bar 
to examine the current demonstration.
The blurred trail of the current trajectory (\textit{ghosted}) demonstration is
also shown in the 3-D viewer to show its full expected path.


Generating a full trajectory from scratch can be difficult for non-experts. 
Thus, similar to  \citet{forbes2014robot}, we provide a trajectory that
the system has already seen for another object as the initial starting trajectory to 
edit.\footnote{We have made sure that it does not initialize with trajectories from other
folds to keep \textit{5-fold cross-validation} in experiment section valid.}

In order to simulate a realistic experience of manipulation,
instead of simply showing a static point-cloud,
we have overlaid CAD models for parts such as `handle'
so that functional parts actually move as the user tries to manipulate the object.

A demonstration can be edited by: 1) modifying the position/orientation
of a waypoint, 2) adding/removing a waypoint,
and 3) opening/closing the gripper.
Once a waypoint is selected, the PR2 gripper is shown with six directional arrows 
and three rings. Arrows are used to modify position
while rings are used to modify the orientation.
To add extra waypoints,
the user can hover the cursor over an  interpolated  (gray) 
waypoint on the \textit{edit bar} and click the plus(+) button.
To remove an existing waypoint, the user can hover over it on the \textit{edit bar}
and click minus(-) to remove.
As modification occurs, the edit bar and ghosted demonstration are updated with a new interpolation.
Finally, for editing the status (open/close) of the gripper, the user can simply click on the gripper.

For broader accessibility, all functionality of Robobarista, 
including 3-D viewer, is built using Javascript and WebGL.


\vspace*{\sectionReduceTop}
\section{Experiments}
\label{sec:experiments}
\vspace*{\sectionReduceBot}


\noindent
\textbf{Data.}
In order to test our model,
we have collected a dataset of 116 point-clouds of objects
with 249 object parts 
(examples shown in Figure~\ref{fig:data_ex}).
There are also a total of 250 natural language
instructions (in 155 manuals).\footnote{Although not necessary for training our model,
 we also collected trajectories from the expert for evaluation purposes.}
Using the crowd-sourcing platform Robobarista,
we collected 1225 trajectories for these objects
from 71 non-expert users on the Amazon Mechanical Turk.
After a user is shown a 20-second instructional video, the user first completes 
a 2-minute tutorial task.
At each session, the user was asked to complete 10 assignments 
where each consists of an object and a manual to be followed.

For each object, we took raw RGB-D images with the Microsoft Kinect sensor and
stitched them using Kinect Fusion \cite{izadi2011kinectfusion} to form a denser point-cloud
in order to incorporate different viewpoints of objects.
Objects range from kitchen appliances such as `stove', `toaster', and `rice cooker' to
`urinal', `soap dispenser', and `sink' in restrooms. 
The dataset will be made available at {\small \url{http://robobarista.cs.cornell.edu}}



%

\noindent
\textbf{Baselines.}
We compared our model against several baselines:

\noindent
1) \textit{Random Transfers (chance)}:
Trajectories are selected at random from the set of trajectories in the training set.

\noindent
2) \textit{Object Part Classifier}:
\jae{To} test our hypothesis that intermediate step of \jae{classifying} object part does not 
guarantee successful transfers, we built an object part classifier using
multiclass SVM \cite{tsochantaridis2004support} on point-cloud features
including local shape features \cite{koppula2011semantic},
histogram of curvatures \cite{Rusu_ICRA2011_PCL}, and distribution of points.
Once classified, the nearest neighbor among the same object part class is selected for transfer.

\noindent
3) \textit{Structured support vector machine (SSVM)}:
\jae{It is a standard practice to hand-code features for SSVM \cite{tsochantaridis2005large}, 
which is solved with the cutting plane method \cite{joachims2009cutting}. 
We used our loss function (Sec.~\ref{sec:metric})  to train and experimented with many state-of-the-art features.}

\noindent
4) \textit{Latent Structured SVM (LSSVM) + kinematic structure}:
The way an object is manipulated depends on its internal structure, 
whether it has a revolute, prismatic, or fixed joint. 
Borrowing from \citet{sturm2011probabilistic},
we encode joint type, center of the joint, and axis of the joint as the latent variable $h \in \mathcal{H}$
in Latent SSVM \cite{yu2009learning}.


\noindent
5) \textit{Task-Similarity Transfers + random}: 
It finds the most similar training task using $(p, l)$
and transfer \jin{any} one of the trajectories from the most similar task.
The pair-wise similarities between the test case
and every task of the training examples
are computed by average mutual point-wise distance \jin{of two point-clouds} after ICP \cite{besl1992method}
and \jf{similarity in bag-of-words representations of language}.

\noindent
6) \textit{Task-similarity Transfers + weighting}:
\jf{The previous method is problematic when non-expert demonstrations for the same task have varying qualities.}
\citet{forbes2014robot} introduces a score function for weighting
demonstrations based on weighted distance to the ``seed'' (expert) demonstration.
Adapting to our scenario of not having any expert demonstration,
we select the $\tau$
that has the lowest average distance from all other demonstrations for the same task
(similar to noise handling of Sec.~\ref{sec:deep}).


\noindent
7) \textit{Our model without Multi-modal Layer}:
This deep learning model concatenates all the input of three modalities
and learns three hidden layers before the final layer. 

\noindent
8) \textit{Our model without Noise Handling}:
Our model is trained without noise handling.
All of the trajectory collected from the crowd was trusted
as a ground-truth label.

\noindent
9) \textit{Our model with Experts}:
Our model is trained using \jin{trajectory demonstrations from an expert} which were collected
for evaluation purpose.



\begin{wraptable}{r}{7cm}

\begin{center}
\vskip -.4in
\caption{
\textbf{Results on our dataset} with \emph{5-fold cross-validation}. 
Rows list models we tested including our model and baselines.
And each column shows a different metric used to evaluate the models.
}
\scriptsize
\begin{tabular}{@{}r|c|c|c@{}}
\hline
 &  \textbf{per manual} & \multicolumn{2}{c}{\textbf{per instruction}} \\ \hline
\textbf{Models}  &  \textbf{DTW-MT}  & \textbf{DTW-MT}  & \textbf{Accuracy ($\%$)} \\ \hline
\emph{chance}   & $28.0 \;(\pm 0.8)$ & $27.8 \;(\pm 0.6)$ & $11.2 \;(\pm 1.0)$ \\ \hline
\emph{object part classifier}      & - & $22.9 \;(\pm 2.2)$ & $23.3 \;(\pm 5.1)$ \\ 

\emph{Structured SVM}              & $21.0 \;(\pm 1.6)$ & $21.4 \;(\pm 1.6)$ & $26.9 \;(\pm 2.6)$ \\ 

\emph{LSSVM + kinematic} \cite{sturm2011probabilistic}    & $17.4 \;(\pm 0.9)$ & $17.5 \;(\pm 1.6)$ & $40.8 \;(\pm 2.5)$\\ \hline

\emph{similarity + random}                           & $14.4 \;(\pm 1.5)$ & $13.5 \;(\pm 1.4)$ & $49.4 \;(\pm 3.9)$   \\ 

\emph{similarity + weights} \cite{forbes2014robot}  & $13.3 \;(\pm 1.2)$ & $12.5 \;(\pm 1.2)$ & $53.7 \;(\pm 5.8)$     \\ \hline

\emph{Ours w/o Multi-modal}  & $13.7 \;(\pm 1.6)$ &  $13.3 \;(\pm 1.6)$ &  $51.9 \;(\pm 7.9)$  \\
\emph{Ours w/o Noise-handling}      & $14.0 \;(\pm 2.3)$ &  $13.7 \;(\pm 2.1)$ &  $49.7 \;(\pm 10.0)$    \\
\emph{Ours with Experts}            & $12.5 \;(\pm 1.5)$ &  $12.1 \;(\pm 1.6)$ &  $53.1 \;(\pm 7.6)$   \\ \hline
\textbf{\emph{Our Model}}                & $13.0 \;(\pm 1.3)$ &  $12.2 \;(\pm 1.1)$ &  $\textbf{60.0} \;(\pm 5.1)$  \\ \hline
\end{tabular}
\label{tab:results}
\end{center}
\vskip -.25in

\end{wraptable}

\vspace*{\subsectionReduceTop}
\subsection{Results and Discussions}
\vspace*{\subsectionReduceBot}

We evaluated all models on our dataset using \emph{5-fold cross-validation}
and the \jae{results are} in Table~\ref{tab:results}. 
Rows list the models we tested including our model and baselines. Each column
shows one of three evaluations. First two use dynamic time warping for
manipulation trajectory (DTW-MT) from Sec.~\ref{sec:metric}.
The first column shows averaged DTW-MT for each instruction manual 
consisting of one
or more language instructions. The second column shows averaged DTW-MT for every
test pair $(p,l)$.

As DTW-MT values are not intuitive, we added the extra column ``accuracy'',
which shows the percentage of \jf{transferred} trajectories with DTW-MT value less than $10$.
Through expert surveys, we found that when DTW-MT of manipulation trajectory
is less than $10$, 
the robot came up with a reasonable trajectory and will very likely be able to accomplish
the given task.

\noindent
\textbf{Can manipulation trajectory be transferred from completely different objects?}
Our full model performed $60.0\%$ in accuracy (Table~\ref{tab:results}),
outperforming the chance as well as other baseline algorithms we tested on our dataset.

\begin{figure*}[tb]
\begin{center}
\includegraphics[width=\textwidth]{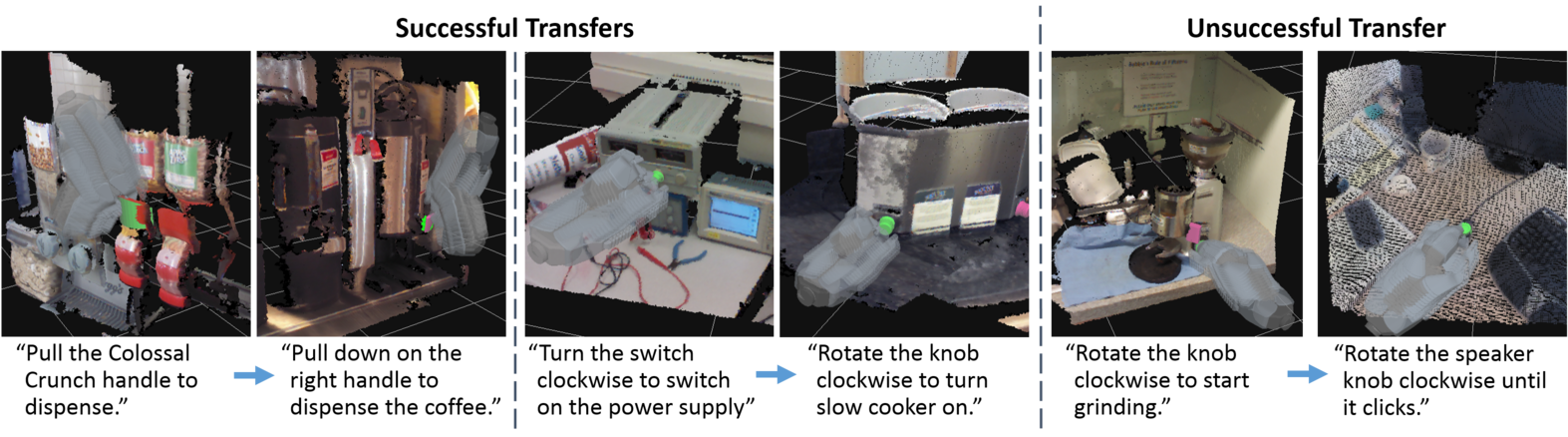}
\end{center}
\vskip -0.2in
\caption{
\textbf{Examples of successful and unsuccessful transfers} of manipulation trajectory from left to right
using our model.
In first two examples, though the robot has never seen  the `coffee dispenser' and `slow cooker' before, 
the robot has correctly identified that the trajectories of `cereal dispenser' and `DC power supply',
respectively, can be used to manipulate them.
}
\label{fig:success_trans}
\vskip -.1in
\end{figure*}

Fig.~\ref{fig:success_trans} shows two examples of successful transfers and one 
unsuccessful transfer by our model.
In the first example, the trajectory for pulling down on a cereal dispenser
is transferred to a coffee dispenser.
Because our approach to trajectory representation is based on the principal axis
 (Sec.~\ref{sec:transfer_adapt}),
even though cereal and coffee dispenser handles are located and oriented differently, the transfer is a success.
The second example shows a successful transfer from a DC power supply to a slow
cooker, which have ``knobs'' of similar shape. The transfer was successful despite
the difference in instructions (``Turn the switch..'' and ``Rotate the knob..'')
and object type.

The last example of Fig.~\ref{fig:success_trans} shows an unsuccessful
transfer. Despite the similarity in two instructions, transfer was unsuccessful because
the grinder's knob was facing towards the front and the speaker's knob was facing upwards.
We fixed the $z$-axis along gravity because point-clouds are noisy and gravity 
can affect some manipulation tasks,
but a more reliable method for finding the object coordinate frame and a 
better 3-D sensor should allow for more accurate transfers.


\begin{figure*}[t]
\begin{center}
\includegraphics[width=\textwidth]{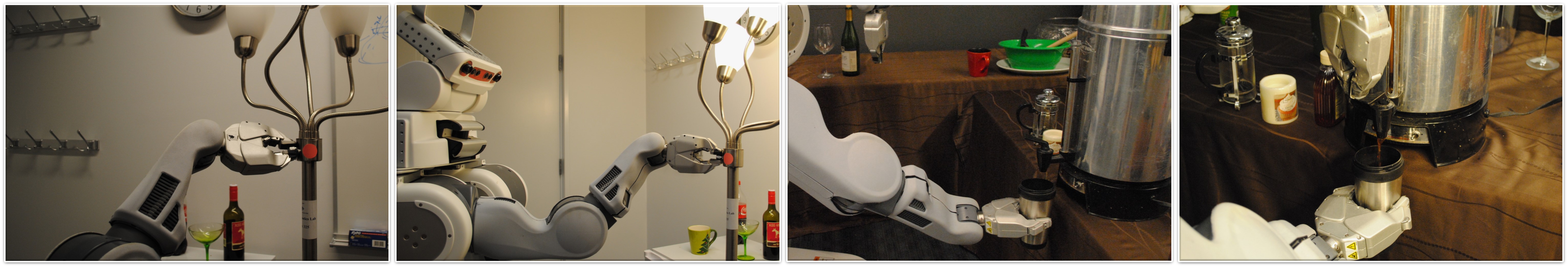}
\end{center}
\vskip -.22in
\caption{
\textbf{Examples of transferred trajectories} being executed on PR2.
On the left, PR2 is able to rotate the `knob' to turn the lamp on.
On the right, using two transferred trajectories, PR2 is able to hold the cup below the `nozzle'
and press the `lever' of `coffee dispenser'. 
}
\label{fig:robotic_exp}
\vskip -.1in
\end{figure*}

\noindent
\textbf{Does it ensure \jf{that the} object is actually correctly manipulated?}
We do not claim that our model can find and execute manipulation trajectories
for all objects. 
However, for a large fraction of objects which the robot has never seen before,
our model outperforms other models in finding correct manipulation trajectories.
The contribution of this work is in the novel approach to manipulation planning
which enables robots to manipulate objects they have never seen before.
\jin{\jae{For some of the objects,} correctly executing a transferred manipulation trajectory may require incorporating visual
and force feedbacks \cite{wieland2009combining,vina2013predicting} \jae{in order for the execution} to adapt exactly to the object
as well as find a collision-free path \cite{stilman2007task}.}

\noindent
\textbf{Can we crowd-source the teaching of manipulation trajectories?}
When we trained our full model with expert demonstrations, which were collected for evaluation purposes, 
it performed \jf{at} $53.1\%$ compared to $60.0\%$ \jf{by our model} trained with crowd-sourced data.
Even with the significant noise in the label as \jin{shown} in last two examples of Fig.~\ref{fig:data_ex},
we believe that our model with crowd demonstrations performed better
because our model can handle noise and because deep learning benefits from having a larger amount of data.
Also note that all of our crowd users are real non-expert users from Amazon Mechanical Turk.

\noindent
\textbf{Is segmentation required for the system?}
In vision community, even with the state-of-the-art techniques \cite{felzenszwalb2010object,krizhevsky2012imagenet}, 
detection of `manipulatable' object parts such as `handle' and `lever' in point-cloud is by itself 
a challenging problem \cite{lai_icra14}.
Thus, we rely on human expert to pre-label parts of object to be manipulated.
The point-cloud of the scene is over-segmented into thousands of supervoxels,
from which the expert chooses the part of the object to be manipulated.
\jin{Even with the \jf{input of the} expert, segmented point-clouds are still extremely noisy
because of the \jae{poor performance of the sensor} on object parts with glossy surfaces.}

\noindent
\textbf{Is intermediate object part labeling necessary?}
The \emph{Object Part Classifier} performed at $23.3\%$,
even though the multiclass SVM for finding object part \jf{label}
achieved over $70\%$ accuracy in five major classes of object parts
(`button', `knob', `handle', `nozzle', `lever') among 13 classes.
\jin{Finding the part label is not sufficient for finding
a good manipulation trajectory because of large variations.}
Thus, our model which does not need \jf{part labels} outperforms the \emph{Object Part Classifier}.

\noindent
\textbf{Can features be hand-coded? What kinds of features did the network learn?}
For both SSVM and LSSVM models, we experimented with several state-of-the-art features for many months, and
\jin{they} gave $40.8\%$.
The \textit{task similarity} method gave a better result of $53.7\%$, but 
it requires access to all of the raw training data (all point-clouds and language) at test time,
which leads to heavy computation at test time and requires a large storage 
as the size of training data increases.

\begin{figure}[tb]
\begin{center}
\includegraphics[width=0.9\columnwidth]{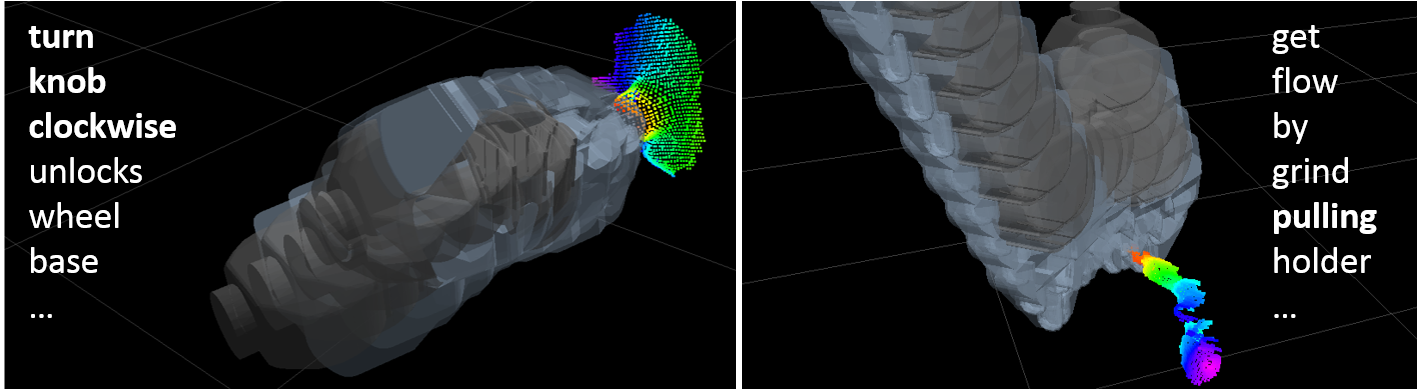}
\end{center}
\vskip -0.2in
\caption{
\textbf{Visualization} of a sample of learned high-level feature (two nodes) at last hidden layer $h^3$.
The point-cloud in the picture is given arbitrary axis-based color for visualization purpose.
The left shows a node $\#1$ at layer $h^3$  that learned to (``turn'', ``knob'', ``clockwise'') along with relevant
point-cloud and trajectory. The right shows a node $\#51$ at layer $h^3$ that learned to ``pull'' handle.
The visualization is created by selecting a set of words, a point-cloud, \jf{and} a trajectory that \jf{maximize} the activation
at each layer and passing the highest activated set of inputs to higher level.
}
\label{fig:learned_rep}
\vskip -.2in
\end{figure}

While it is extremely difficult to find a good set of features for three modalities,
our deep learning model which does not require hand-designing of features
learned features at the top layer $h^3$  
such as \jin{those} shown in Fig.~\ref{fig:learned_rep}.
The left shows a node that correctly associated point-cloud (axis-based coloring), trajectory,
and language for the \jin{motion} of turning a knob clockwise.
\jin{The right shows a node that correctly associated for the motion of pulling the handle.}

Also, as shown for two other baselines using deep learning,
when modalities were simply concatenated, it gave $51.9\%$,
and when noisy labels were not handled, it \jf{gave only} $49.7\%$.
Both results show that our model can handle noise from crowd-sourcing
while learning relations between three modalities.

\vspace*{\subsectionReduceTop}
\subsection{Robotic Experiments} 
\vspace*{\subsectionReduceBot}
As the PR2 robot stands in front of the object,
the robot is given a natural language instruction 
and segmented point-cloud.
\jf{Using our algorithm, manipulation trajectories to be transferred were found for the given point-clouds
and languages.}
Given the trajectories which are defined as set of waypoints, the robot followed
the trajectory by impedance controller (\texttt{ee\_cart\_imped}) \cite{bollini2011bakebot}.
Some of the examples of successful execution on PR2 robot are shown in Figure~\ref{fig:robotic_exp}
and in video at the project website: 
{\small \url{http://robobarista.cs.cornell.edu}}

\vspace*{\sectionReduceTop}
\section{Conclusion}
\vspace*{\sectionReduceBot}

In this work, we  introduced a novel approach to predicting manipulation trajectories via part based transfer,
which allowed robots to successfully manipulate objects it has never seen before.
We formulated it as a structured-output problem and 
presented a deep learning model capable of handling three completely different modalities
of point-cloud, language, and trajectory while dealing with large noise in the manipulation demonstrations.
We also designed a crowd-sourcing platform Robobarista 
that allowed non-expert users to easily give manipulation demonstration over the web.
Our deep learning model was evaluated against many baselines
on a large dataset of 249 object parts with 1225 crowd-sourced demonstrations.
In future work, we plan to share the learned model using the knowledge-engine, RoboBrain \cite{saxena_robobrain2014}.

\vskip -2in
\subsubsection*{Acknowledgments}
\vspace*{\sectionReduceBot}

We thank Joshua Reichler for building the initial prototype of the crowd-sourcing platform.
We thank Ian Lenz and Ross Knepper for useful discussions.
This research was funded in part by Microsoft Faculty Fellowship (to Saxena), NSF Career award (to Saxena) and Army Research Office.

{
\renewcommand\refname{\vskip -1.5cm} 
\scriptsize
\bibliography{writeup}
\bibliographystyle{abbrvnat}
}


\newpage
\clearpage
 

\end{document}